# E-Mail Assistant – Automation of E-Mail Handling and Management using Robotic Process Automation

Arpit Khare
*Department of Electronics and Communication*
*University of Allahabad*
Prayagraj, India
arpitkhare33@gmail.com

Sudhakar Singh
*Department of Electronics and Communication*
*University of Allahabad*
Prayagraj, India
sudhakar@allduniv.ac.in

Richa Mishra
*Department of Electronics and Communication*
*University of Allahabad*
Prayagraj, India
richa_mishra@allduniv.ac.in

Shiv Prakash
*Department of Electronics and Communication*
*University of Allahabad*
Prayagraj, India
shivprakash@allduniv.ac.in

Pratibha Dixit
*King George's Medical University*
Lucknow, India
pdkgmu@gmail.com

*Abstract*—In this paper, a workflow for designing a bot using Robotic Process Automation (RPA), associated with Artificial Intelligence (AI) that is used for information extraction, classification, etc., is proposed. The bot is equipped with many features that make email handling a stress-free job. It automatically login into the mailbox through secured channels, distinguishes between the useful and not useful emails, classifies the emails into different labels, downloads the attached files, creates different directories, and stores the downloaded files into relevant directories. It moves the not useful emails into the trash. Further, the bot can also be trained to rename the attached files with the names of the sender/applicant in case of a job application for the sake of convenience. The bot is designed and tested using the UiPath tool to improve the performance of the system. The paper also discusses the further possible functionalities that can be added on to the bot.

*Keywords—E-Mail Automation, Artificial Intelligence (AI), Robotic Process Automation (RPA), Expert Systems, Virtual Assistant, UiPath.*

## I. INTRODUCTION

Electronic mail also well-known as email or e-mail since 1993 is basically used to exchange digital information among recipients [1]. The present email technology has three parts, the message envelope, the message header, and the message body [2]. The common web-based email services include Gmail, Outlook, Hotmail, Yahoo! Mail, etc. [3]. As per the consumer email survey conducted on Adobe's consumers in 2018, office workers check their electronic mail account on an average of 2.5 hours each day [4]. When it comes to work-related e-mails, the average human being spends more than 3 hours a day. That is a huge amount of time spent on repetitive communication instead of actual creative work [4]. Emails are the leading element in all official communications with nearly 86% of business professionals choosing emails as their preferred mode of communication [4]. As per a survey [4], 90% of consumers check their email during office hours, while 9% check their emails constantly. Also, 85% of consumers check their email before getting to work, and more than 25% do so as soon as they wake up on the bed [4].

Considering the situation of a business owner, it is quite a tedious task to handle the jobs/internship application performed through email communication. Therefore, automation in this service is of utmost essential so that the working population in the world can be free from the irrelevant workload of managing their mailboxes. Hereby, we propose the approach for building an E-mail Assistant that is adept at a lot of tasks related to streamlining a person's email while he/she works on the other important aspects of running his/her business. It stops wasting time sorting through the inbox, addition, deletion, etc. Further, these problems fall under the category of finding similar substructures problem which is an NP-complete problem [5] in nature and needs an efficient algorithm to solve the exponential order problem.

The emergence of artificial intelligence (AI) in the former era has changed the process of handling business related to the use of machines instead of a human being hence using robotic process automation (RPA) [6]. This paradigm shift not only supports organizations but also increases profit by reducing waste. Likewise, new customer registration requires several checks and information. This whole process can be changed into a series of tasks where robot system can be trained and repeated over the relevant systems [7]. So now this whole process can be managed by the robots for 24x7 perfectly and deprived of any error. This technology is different from other sectors, frequent repetition of the process can be treated operation-intensive, where employees are not needed to take initiative, are subjective to a rule with already established standards [6].

The proposed bot is designed using UiPath [8], a robotic process automation (RPA) tool [9]. It is equipped with the following features to make email handling a stress-free job. It automatically login into the mailbox through secured channels and distinguishes between useful and not useful emails based on the provided keywords. It classifies the mails into different directories e.g., Work (for work-related emails, CV, Resumes) and Receipts (for Bills and Invoices). It downloads the attachments (if any) from the classified emails, creates different directories for different categories, and stores the downloaded files into the relevant directories. It moves the not useful mails into the trash. This bot can also be trained to rename the attachment files with the names of the sender/applicant in case of a job application for the sake of convenience. The bot is tested on several test emails to verify its proper functioning.

The rest of the paper is organized as follows. Section two consists of a review of related works. Section three describes the design of the proposed E-Mail Assistant. Testing of the designed bot is carried out in Section four. Finally, Section five concludes the paper with the scope of future works.

## II. RELATED LITERATURE

Presently advanced technologies transformed the business process into automated systems using the software robots which are based on predefined algorithms by using AI. It can also be referred to as a digital worker and broadly termed as "robotic process automation" which leads to organizational and technological change. It includes specific tool or software (including cognitive software), eventually leading to the foundation of a hybrid workforce [10].

Several techniques have been developed for automating E-Mail system and enabling it with smart features. Email spam, also known as junk email is one of the most tedious issues. Many times valid and relevant emails are received in the spam [11]. The huge number of incoming emails is another problem, users waste a lot of time and energy in the efforts of identifying relevant and useful emails. To avoid this problem, emails need to be categorized and labeled based on the inside information, so that users can identify the useful emails even before opening them [11]. Alsmadi and Alhami [11] proposed a machine learning-based solution to predict the real sender of a mail by training the model using some training mails from the past. This approach provided a good accuracy to the prediction but still, the identification of real sender was trapped in the probabilities of machine learning models. The uncertainties of identification cannot be implemented in real-life business problems, as it could lead to some hazardous situations for a business owner. The amount of work performed to just identify the sender is also not feasible. Thus, there is a requirement of a technique to perform this task with full accuracy and the least effort.

The hypothesis of the study is that the Case Based Reasoning (CBR) approach can be employed to solve the problem of email overload and this can be analyzed by investigating the email data sets. It shows that it mapped the forthcoming queries by using analogous prior queries and reprocessing the responses [12]. Email management requires significant efforts from both senders and recipients which enabled the automation of email processing. The hybrid techniques are required to study what automation is required by users [13]. This study helps us to understand novel end-to-end systems for generation of short email responses automatically [14]. Moreover, it also gives an idea on RPA which is a software-based solution completely. For instance imitating human activities in a sequential way that leads to meaningful action, without any kind of human intervention is referred as Robotic Process Automation [15]. It can also provide important data on email classification along with utilizing natural language processing and data mining activities, Spam detection, etc. [11]. Furthermore, the social behaviors of the users are used to determine a novel email classification method for enterprises.

Automatic classification of PDF text is a substantial problem in the E-Mail system. Bui et al. [16] have addressed the classification of PDF text to extract the information from publications or reports using Information Extraction (IE) systems. Authors designed a text classification method that automatically classifies PDF text into title, abstract, main content, semi-structured, and metadata categories. Automatic classification of email messages into different folders/directories is an open and challenging problem, particularly for the machine learning algorithms. Tam et al. [17] used supervised learning algorithms for organizing emails into different folders automatically. They discussed the problems due to the different semantics applied by the users. The meaning of the different folders varies from person to person which is an obstacle for learning methods. Further, as the number of emails being received is increasing, efficient automatic foldering is getting essential [18]. Brutlag and Meek [19] were the first to analyse email classification as the problem of text classification. An extensive study was carried out in [20] on the email foldering problem using the Enron Corpus [21] dataset.

Dredze et al. [22] proposed the idea of intelligent email by applying AI to email. They considered the user-oriented approach and applied the concepts of machine learning and NLP (natural language processing) to propose intelligent email and defined it as an intelligent system for supporting email interfaces. Dalli et al. [23] designed an Adaptive Information Management (AIM) service to be used in a voice-based Virtual Personal Assistant (VPA). The AIM service is consisting of three components: an email summariser, email categoriser, calendar scheduling, and an adaptive prioritisation service.

The proposed approach uses similar functionalities as discussed above to integrate into one software robot using RPA. The frequency of emails per day even in a mid-size organization is large enough and needs a storage of huge size. The recent technological developments can manage big data [24] so that the massive email traffic can be handled efficiently. Segregation of emails can be done into different groups and the responses can be sent by the RPA based solution while the critical ones not assigned in any group can be controlled by the particular personnel. Increased workload on users demands more efficient works for responding the query. There are other aspects also which need to be recognized like tagging and prioritizing of incoming emails, moving these emails to other locations, its labeling and, sending emails at scheduled time or in a particular context, also to the right recipient. RPA is an evolving technique that uses software and algorithmic programmed systems which act as humans for the support of proficient business processes. It leads to a decrease in the cost of human resource-related spending by 20–50 percent and transaction processing costs by 30–60 percent [25].

RPA is highly used in industry but also an attractive field of research in the scientific domain. Ribeiro et al. [26] investigated the contribution of RPA tools with AI for the improvement of ERP-related processes. Authors in [27] have focused on the gap between the virtual assistant and recommendation system in the view of various technological aspects for designing a conversational system. Enriquez et al. [28] presented a systematic mapping study on RPA based research and development in the scientific and industrial domain. A systematic literature review is carried out in [9] on RPA. An assessment framework derived from this literature review is also presented. All these works show the importance and pertinence of RPA which needs to be explored and applied in the different domains to design the efficient bot.

The bot has become a commonly used automation system for large-scale scientific computing besides data-intensive applications. Each bot is typically linked with a deadline of execution to guarantee the Quality of Service (QoS) such as security, flow-time, reliability, and trust, etc.; and low QoS typically imposes penalties, however, the provider of services may charge users based on user demand and QoS parameters obtained [29]. Further, the applications of the bot in real-life problems [26]–[28] are such as the loop bounds are known during compile-time, data-flow computation, and various numerical algorithms e.g. Gaussian elimination, Fourier transform, and its variants. Moreover, RPA which is used in bot has many benefits in automation of organization and business methods associated with advances of artificial intelligence methods, algorithms, tools, and techniques which allows improved accuracy, precision, and execution of RPA methods to extract the information in classification, regression, recognition forecast, and optimization.

III. PROPOSED DESIGN OF THE E-MAIL ASSISTANT

The E-Mail Assistant would have the following functionalities and features.

- A secured robot that could login to the system when allowed to do so.
- Identify the senders without manually checking on devices.
- Save the user's time by not opening the mails from advertising agencies/companies.
- Increase the user's convenience by classifying the mails into various categories.
- Download the attachments if any automatically without human intervention.
- Classify the downloaded attachments into two folders – Useful and Not Useful.
- Inside the Useful folder, further classify them into various folders like – bills, resume, offer letter, invoices, etc.

Fig. 1 depicts the system design of the proposed E-Mail Assistant. It also shows the order of different activities performed by the bot. The bot is designed and implemented using the UiPath tool on Window's operating system.

UiPath is an RPA platform for the end-to-end high-scale automation. This tool offers solutions for the enterprises to automate the repetitive tasks of offices for rapid business transformation. Using it, one can convert tedious and annoying tasks into an automation process using multiple utilities [8]. The main components of these tools are (i) a native email application server, (ii) an email processor engine, and (iii) an analyzing module. It works effectively due to its independent and integrated issues. The main part is to categorize the incoming emails along with its segregation and generate an autoreply to all the important emails.

*1) Native Email Application Server:* It has a public inbox that receives incoming emails from exterior to the environment. These emails are referred to an Email Logger.

*2) Email Processor Engine:* This engine comprises of the Email Logger, Ingestion Layer, Web Scraping, Data Enrichment Module, RPA Algorithm, and Segregated Mail Router. All of these modules execute in an integrated fashion which tends to attain the envisioned task of segregating the emails.

*3) Email Logger:* It records each and every email with its lot of information (sender, receiver, timestamp) along with its batch ID respectively.

*4) Ingestion Layer:* This layer constitutes the emails present in the email logger which consist of a header, subject, and body i.e. content. It has a unique ID for each email.

*5) Web Scraping:* It deals with the extraction of email body content which preserves the unique ID. It appends the other information related to email. Information stored in this data format is allowed and prepared for analysis in the analyzing module.

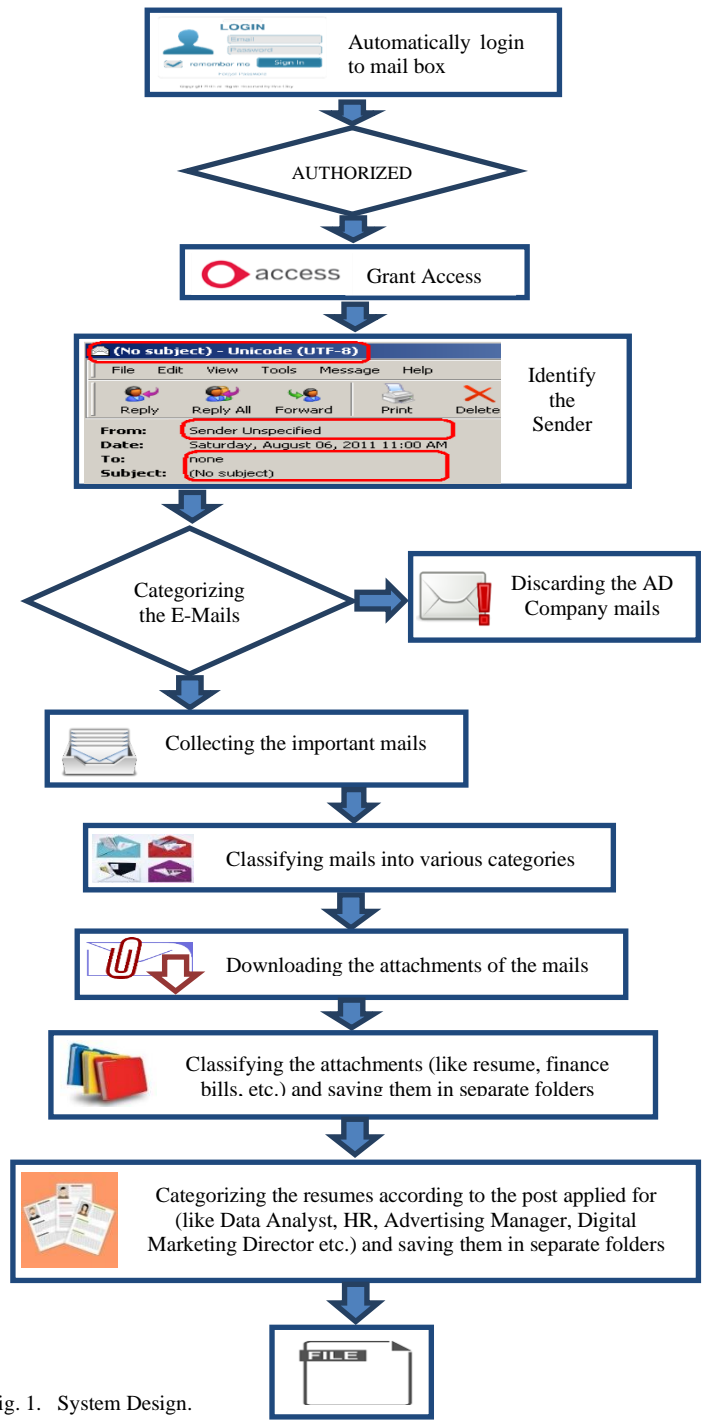

Fig. 1. System Design.

*6) Data Enrichment Module:* This module deals with the removal of irrelevant and noisy features from the emails that optimizes the size of the emails.

*7) RPA Algorithm:* It comprises a protocol with email services in which IDs of email class labels are assigned for every email to direct them respectively.

*8) Segregated Mail Router:* This module directs the autoreply to the respective email ids. This system segregates the inbound email into a local folder according to the topic of discussion and decisive action can be taken further. It also consists of a prediction module that does feature selection and its representation of the content of the emails which can predict the probable responses to the emails one by one using the permitted response and cluster.

There are many email activities available in UiPath software and each one is based on the different protocols to be used like Internet Message Access Protocol (IMAP), Post Office Protocol version 3 (POP3), Simple Mail Transfer Protocol (SMTP), Exchange, and Outlook. In the proposed bot, IMAP is used. In the UiPath, the "GetIMAPMailMessages" mail message activity is configured first. It is configured with the Mail Folder name, Port number, Server name, email-id, and password. Generally, the mail folder name is "Inbox". If the robot is being configured with a Gmail account, then port number 993 is used in most cases. For a Gmail account, the server's name is "imap.gmail.com". Email-id and password can be provided through the window's credential manager [30]. It is also possible to set the number of emails which is executed by the robot in one go. The output argument of this "GetIMAPMailMessages" mail messages activity is a List<MailMessage>. This output can be stored in any variable, say "output_mail". Further, the list of unwanted mail senders which are already known can be also provided through an excel sheet, so the mails received from those will be considered not useful.

Emails are classified into various categories like bills/invoices, resumes, etc. on the basis of some keywords. To download and save the attachment at the destined folder, the Save attachments activity is used and provided with the Folder Path i.e., the path of the folder where the attachments are intended to get downloaded, and also provided with each mail from the List<MailMessage>. The robot is trained to open each downloaded file one by one and recheck the information given in it. It identifies the useful attachments. The definition of useful attachments can be modified as per the user's requirements. In this paper, the following definition is used – If any document contains the following keywords: resume, cv, bill, etc. it is classified to be useful. Further, the useful attachments are classified into various folders as per the above keywords. The windows activities can be used to train the robot to rename the attachment files to appropriate names if the application demands to do so in the real world. The various parameters used in the activity are shown as screenshot in Fig. 2 and described in TABLE I.

The procedure discussed above to design the proposed E-Mail Assistant can be represented in the form of the algorithm as shown in Algorithm 1.

## IV. WORKFLOW EXECUTION AND TESTING

The UiPath version-2019.10.1 (free edition) was installed on the Windows 10 operating system running on the machine with 64-bit Intel(R) Core(TM) i5-8250U CPU @ 1.60GHz and 8 GB RAM. The objective of testing is to demonstrate that the bot is working properly with the functionalities specified.

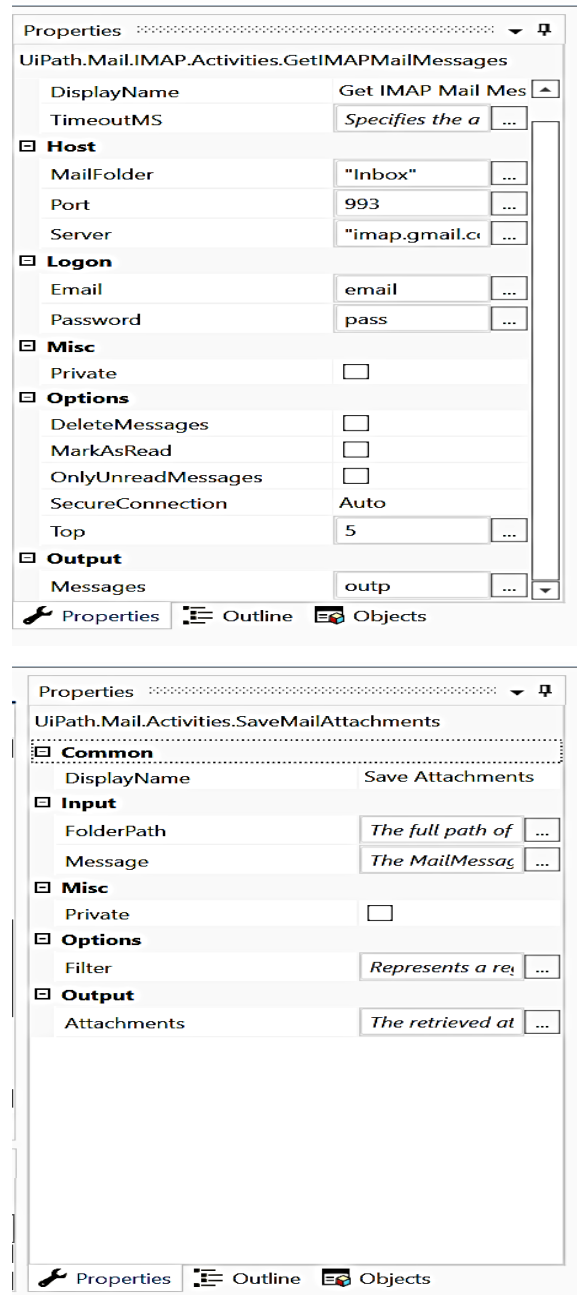

Fig. 2. Configuration of "GetIMAPMailMessages" activity

TABLE I. PARAMETERS USED IN THE ACTIVITY

| Parameters | Description |
|---|---|
| TimeoutMS | The amount of time (in milliseconds) activity should be allowed to wait before giving an error message, in case of any delay. |
| MailFolder | The folder of Gmail should be the working directory for the bot. |
| Port | The port number on which the robot should receive the emails. |
| Server | The server's name from which the emails should be extracted ("imap.gmail.com"). |
| Email | The email id of the user can be given at the run time through Window's Credential Manager. |
| Password | The password for the email id of the user can be given at the run time through Window's Credential manager. |
| Top | The number of emails that should be taken into work (starting from the top). |
| Messages | The variable in which all the messages are stored. |

---

**Algorithm 1. Algorithm of proposed E-Mail Assistant**

Configure the Get IMAP mail message activity
Give permission to the robot from the google account by "Allowing access of less secure apps"
Configure the Window's credential manager with email-id and password
output_mail := List<MailMessage>
For each mail-in output_mail repeat
    if is_important(mail)
        open the mail
  Classify the mail into various categories based on some keywords
  Download the attachment file
  Classify the attachment into important and non-important
  Classify the important attachments into folders: Bills, Resumes, Invoices, etc.
  Classify the Resume files, into eligible and not eligible candidates
  Rename all of them with the sender's name, designation, and date of application
  Sending a mail to all eligible candidates to call for an interview (optional)
End

---

Several Gmail account holders were requested to send emails of different types with subjects containing keywords like "bills", "invoice", "role" etc. Each such email also contains an attachment of miscellaneous types. As soon as the bot is being executed, it starts performing classification of the mails on the basis of the above-defined keywords into two folders/labels of Gmail i.e., Work and Receipt. Also, the bot creates different folders in the local machine for downloading the files attached with the incoming mails, and thus downloads the attachments into the two different folders Work and Receipt. The screenshot of the main workflow with the different components of this bot is shown in Fig. 3.

The inbox page of the test email-id received the emails which are from various contacts and of various categories (e.g.- Receipts, Resume, Bills, etc.). All these incoming emails are the input to the test cases. In the next step of workflow execution, the bot is allowed to handle all the unread emails of the inbox. The bot is provided with the following instructions: move the emails with the keyword "Resume" in the subject to the folder/label "Work", similarly, the emails with the keyword "Bill" or "Invoice" to the folder/label "Receipt". Note that, these emails are not automatically categorized by the classification algorithms of Google in Gmail. Instead, it is performed on the basis of the instructions given to the bot allowing it to customize the classification. Also, the emails which are considered to be not important for the user are automatically sent to the trash. We can extend this classification to other categories like Spam or any other desired label. Meanwhile, the emails may contain a few attachments. The bot makes the user's task easier and makes them free from opening the emails exclusively. It downloads the attachments and then classifies them into many folders based on the keywords by doing altogether these tasks repeatedly. At this time, the bot makes these folders and then stores the downloaded attachments in the appropriate folders. The Control via different criteria is important for the users to control the bot using different implementations.

*A. Performance Evaluation*

Various QoS (Quality of Service) parameters for the effectiveness of the Email Automation Robot are measured as follows. *Efficiency:* The ratio of valuable work performed by the bot. *Time consumption:* Time taken by the bot to process and throughput. *Accuracy:* A state of the bot to be precise. *Precision:* The Precision indicates how precise (accurate) the model of the bot is.

To assess the efficacy of the system, a Gmail account has been used wherein the incoming emails were read and replied to using two cases.

*1) CASE I – Email processing without the robot (i.e. manually):* In the study published in Harvard Business Review [31], it was observed that an average American spends approximately 78 seconds on an email. This sums up to taking 702 seconds to process 9 emails.

*2) CASE II – E-mail processing with the robot (i.e. automation):* Tasks like opening the email by user's credentials, looking over the received email, choosing important emails, reading the email's content, and responding to the emails personally are done by hand in Case I while in Case II, the bot does the same for the user deprived of any kind of human involvement. It has been found that it is tough for a user to read and segregate all the emails incoming into the inbox, also classify the unwanted and annoying emails, and respond to all the important emails separately. Nonetheless when the user uses the email automation, in which the email is segregated and categorized specifically according to the list automatically when it gets entered into the inbox. It can be efficiently compared to the first case deprived of the automation which requires the least involvement of the user for the work to be performed manually.

*3) Time Consumption:* With reference to Case I, it has been observed that manual work increases in absence of automation which enhances time consumption. At the same time, there is a need to read all the emails, segregate and respond accordingly. Case II is significantly less time-consuming in comparison to the process done without RPA. There are certain factors like the number of receiving emails, the extent of the contents present in subject and body, and size of the attached files that play a key role which determine how much time will be consumed in completing the task in both cases. In Fig. 4, we can see that it took our bot 75 seconds to handle 9 emails with attachments and in classifying the attachments, whereas for an average human being it took 702 seconds to perform the same task.

*4) Accuracy:* The accuracy level is determined by how many emails are appropriately classified in the designated directory/folder for that email. In Case I, when the classification process is done by a human being, it may lead to some mistakes due to human errors in reading or understanding the class of the email, this makes the accuracy of the process to be reduced. While, when this process is done by an RPA bot, then the probability of misclassification of an attachment file that is working on the keywords basis is almost nil, this makes Case II's process of RPA bot much more effective.

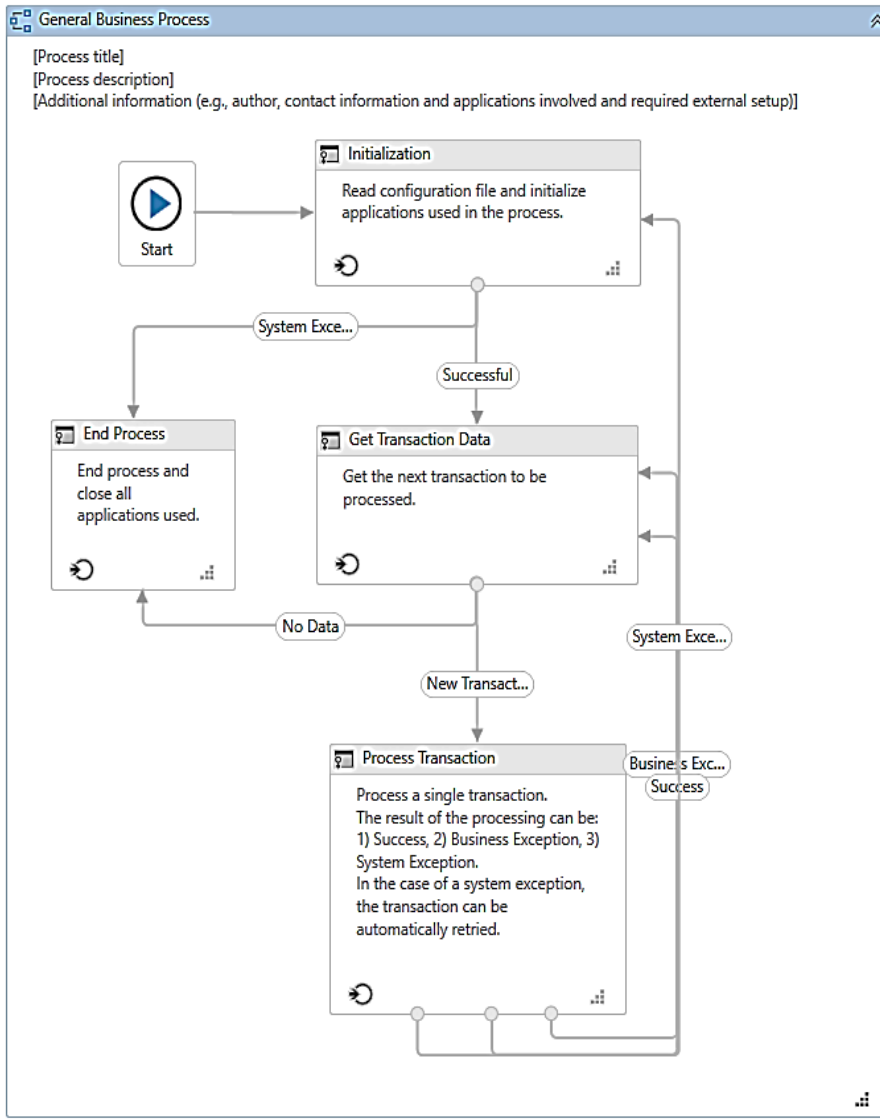

Fig. 3. Workflow of the Robot

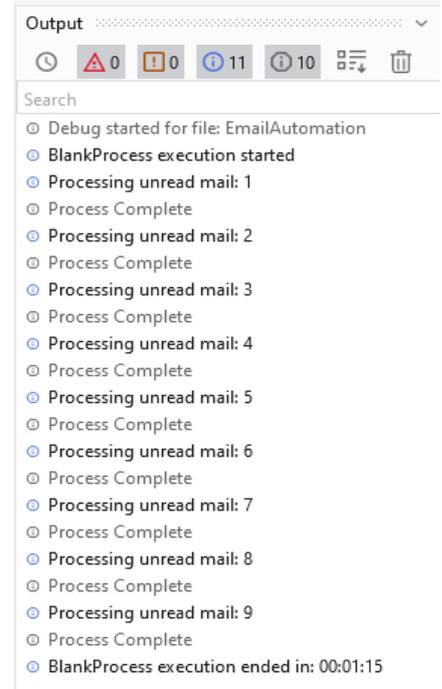

Fig. 4. Execution time for RPA bot to process 9 emails

## V. CONCLUSION AND FUTURE WORK

This paper deals with the automation of E-Mail handling and management. The automation gives relief from logging into the system for inbox management. Regarding this, it proposes a simple and novel workflow to design an Intelligent E-Mail Assistant using UiPath, an RPA tool. The robot provides a good level of comfort to its users with the equipped functionalities. It automatically and securely login into the mailbox, classifies and sub-classifies the emails based on essential keywords. It automatically downloads the attachments and stores them in different folders respective to the classified emails. It can also rename the attached files according to the requirements. The bot does not override human commands/actions to counter the issue of any kind of security threat. The workflow execution and testing have been done on several test emails, and it is found to be working as per the functionalities defined. Using the windows activities, the robot can be trained to rename the attached files of emails to the desired names if the application needs to do so in the real world. For example, it can rename the resumes like <candidate_name>_<highest_qualification>_<application_date> etc. Also, it can send mail to the eligible candidates with an interview call.

Moreover, this bot can be made more intelligent and robust by machine learning approaches for classifying, processing the downloaded images using image processing algorithms, and many more. The bot can be trained to scan the image using an optical character reader (OCR) and to perform classification based on scanning, which will free up the robot from relying on the subject entered by the sender and evaluating the file being sent. Further, it can be also trained to send responses and take actions on the basis of incoming emails. For example, training it to respond to a mail regarding a meeting by checking the schedule in the owner's online calendar, and suggesting some alternate time if the user is already engaged in other meetings during the mentioned time in the received email, will increase the level of comfort of the user. Also, for the emails containing bills and receipts, payment can be done instantly if the user allows the bot to handle this task too, considering the security issues preferred by the user.

On the other hand, one of the key challenges is RPA maintenance, as user interfaces may change more frequently, whereas RPA change, sometimes the RPA reconfiguration is a challenging issue. Therefore, RPA bot optimization will be performed as suggested in the future direction. Further, the scope for UiPath can be extended to imaginable boundaries with different optimization and machine learning techniques.

ACKNOWLEDGMENT

The authors are indebted to the Mae Fah Luang University for supporting the free registration in this conference.